\documentclass[10pt,twocolumn,letterpaper]{article}

\usepackage[table,xcdraw]{xcolor}
\usepackage{3dv}
\usepackage{times}
\usepackage{epsfig}
\usepackage{graphicx}
\usepackage{amsmath}
\usepackage{amssymb}
\usepackage{comment}
\usepackage{makecell}

\usepackage[pagebackref=true,breaklinks=true,colorlinks,bookmarks=false]{hyperref}

\threedvfinalcopy %

\def\threedvPaperID{86} %

\ifthreedvfinal\fi

\definecolor{orange}{rgb}{1,0.5,0}
\definecolor{deeppink}{RGB}{255,20,147}

\newif\ifdraft

\draftfalse
\ifdraft

\newcommand{\TODO}[1]{{\color{blue}{\bf TODO:} #1}}

\else

\newcommand{\TODO}[1]{}

\fi

\begin{document}

\title{Long Term Motion Prediction Using Keyposes}

\author{Sena Kiciroglu$^{1}$ \,\,\, Wei Wang$^{1,2}$ \,\,\,\, Mathieu Salzmann$^{1,3}$ \,\,\, Pascal Fua$^{1}$\\
	$^1$ CVLab, EPFL  \qquad $^2$ MHUG, University of Trento \qquad $^3$ Clearspace  \\
	\tt\small sena.kiciroglu@epfl.ch \vspace*{1.5ex}
}

\maketitle
\thispagestyle{empty}

\begin{abstract}

Long term human motion prediction is essential in safety-critical applications such as human-robot interaction and autonomous driving. In this paper we show that to achieve long term forecasting, predicting human pose at every time instant is unnecessary. Instead, it is more effective to predict a few keyposes and approximate intermediate ones by interpolating the keyposes. 

We demonstrate that our approach enables us to predict realistic motions for up to $5$ seconds in the future, which is far longer than the typical $1$ second encountered in the literature. Furthermore, because we model future keyposes probabilistically, we can generate multiple plausible future motions by sampling at inference time. Over this extended time period, our predictions are more realistic, more diverse and better preserve the motion dynamics than those state-of-the-art methods yield.

\end{abstract}

\vspace{-2mm}
\section{Introduction}

Human motion prediction is a key component of many vision-based applications, such as automated driving~\cite{Habibi18,Fan12b,Shi21}, surveillance~\cite{Morais19,Kiciroglu20}, accident prevention~\cite{Tang20b,Zhao22a}, and human-robot interaction~\cite{Gui18b,Butepage18b}. Its goal is to forecast the future 3D articulated motion of a person given their previous 3D poses. Most approaches formulate this task as one of regressing a person's pose at every future time instant given the past poses. While recurrent neural networks~\cite{Ghosh17,Martinez17b} and graph convolutional networks~\cite{Mao19,Lebailly20,Mao20} are effective for short-term predictions, typically up to one second in the future, their prediction accuracy degrades quickly beyond that, and addressing this shortcoming remains an open problem.

This paper focuses on longer-term prediction, which is critical in many areas, such as providing an autonomous system sufficient time to react to human motions. Our key insight is that, for this task, predicting the pose in \emph{every} future frame is unnecessary. For example, consider a boxing jab motion. The most significant poses are the ones where the hand is closest to the chest and where the arm is the most extended. The in-between poses are transition ones that can be interpolated from these two. Therefore instead of treating a motion as a sequence of consecutive poses, we downsample it to a set of  {\it keyposes} from which all other poses can be interpolated up to a given precision. We then use these keyposes for long-term motion prediction.

The simplest way to so would be to replace the poses in existing frameworks by our keyposes. However, while all keyposes are unique, some tend to be similar to each other. We therefore cluster those we extract from a training set and develop a framework that treats keypose prediction as a classification problem. This has two main advantages. First, it overcomes the tendency of regression-based prediction methods to converge to the mean pose in the long term. Second, it allows us not only to predict the most likely future motion by selecting the most probable clusters but also to generate multiple plausible predictions by sampling the relevant probability distributions. This is useful because people are not entirely predictable, as in the case of a pedestrian standing on the curb who may, or may not, cross the street.

In summary, our contributions are threefold. (i) We introduce a keypose extraction algorithm to represent human motion in a compact way. (ii) We formulate motion prediction as a classification problem and design a framework to predict keypose labels and durations. (iii) We demonstrate that our approach enables us to predict multiple realistic motions for up to $5$ seconds in the future, which is far longer than the typical $1$ second encountered in the literature. The motions we generate preserve the dynamic nature of the observations, whereas the methods designed for shorter timespans tend to degenerate to static poses. Our code and an overview video can be accessed via our project website, \url{https://senakicir.github.io/projects/keyposes}.  %

\section{Related Work}

The complexity of human motion makes deep learning an ideal framework for tackling the task of motion prediction. In this section, we first review the two main classes of deep models that have been used in the field and then discuss approaches that depart from these main trends. Finally, we discuss the use of keyposes for different tasks.

\vspace{-3.5mm}
\paragraph{Human Motion Prediction using RNNs.}

Recurrent neural networks (RNN) are widely used architectures for modeling time-series data, for instance for natural-language processing~\cite{Zhang15d} and music generation~\cite{Sturm16,Simon17c}. Since the work of Fragkiadaki \emph{et al.}~\cite{Fragkiadaki15}, these architectures have become highly popular for human motion forecasting. In this context, the S-RNN of Jain \emph{et al.}~\cite{Jain16} transforms spatio-temporal graphs to a feedforward mixture of RNNs; the Dropout Autoencoder LSTM (DAE-LSTM) of Ghosh \emph{et al.}~\cite{Ghosh17} synthesizes long-term realistic looking motion sequences; the recent Generative Adversarial Imitation Learning (GAIL) of Wang \emph{et al.}~\cite{Wang19h} was employed to train an RNN-based policy generator and critic networks. HP-GAN~\cite{Barsoum18} uses an RNN-based GAN architecture to generate diverse future motions of $30$ frames. 

Despite their success, using RNNs for long-term motion prediction suffers from drawbacks. As shown by Martinez \emph{et al.}~\cite{Martinez17b}, they tend to produce discontinuities at the transition between observed and predicted poses, and often yield predictions that converge to the mean pose of the ground-truth data in the long term. In~\cite{Martinez17b}, this was circumvented by adding a residual connection so that the network only needs to predict the residual motion. Here, we also develop an RNN-based architecture. However, because we treat keypose prediction as a classification task, our approach does not suffer from the accumulated errors that such models tend to generate when employed for regression.

\vspace{-3.5mm}
\paragraph{Human Motion Prediction using GCNs.}

Mao \emph{et al.}~\cite{Mao19} proposed to overcome the weaknesses of RNNs by encoding motion in discrete cosine transform (DCT) space, to model temporal dependencies, and learning the relationships between the different joints via a GCN. Lebailly \emph{et al.}~\cite{Lebailly20} build on top of this work by combining a GCN architecture with a temporal inception layer. The temporal inception layer serves to process the input at different subsequence lengths, so as to exploit both short-term and long-term information. Alternatively, ~\cite{Mao20, Katircioglu22} combine the GCN architecture with an attention module aiming to learn the repetitive motion patterns. These methods constitute the state of the art for motion prediction. Nevertheless, they were designed for forecasting up to 1 second in the future. As will be shown by our experiments, for longer timespans, they tend to degenerate to static predictions.

\vspace{-3.5mm}
\paragraph{Other Human Motion Prediction Approaches.} 

Several other architectures have been proposed for human motion prediction. For example, B{\"u}tepage \emph{et al.}~\cite{Butepage17} employ several fully-connected encoder-decoder models to encode different properties of the data. One of the models is a time-scale convolutional encoder, with different filter sizes. In~\cite{Butepage18b}, a conditional variational autoencoder (CVAE) is used to probabilistically model, predict and generate future motions. This probabilistic approach is extended in~\cite{Butepage19} to  incorporate hierarchical action labels. Aliakbarian \emph{et al.}~\cite{Aliakbarian20} also perform motion generation and prediction by encoding their inputs using a CVAE. They are able to generate diverse motions by randomly sampling and perturbing the conditioning variables. Similarly, Yuan \emph{et al.}~\cite{Yuan20} also use a CVAE based approach to generate multiple futures. Li \emph{et al.}~\cite{Li20b} use a convolutional neural network for motion prediction, producing separate short-term and long-term embeddings. In~\cite{Corona20a,Adeli20}, interactions between humans and objects in the scene are learned for context-aware motion prediction. Aksan \emph{et al.}\cite{Aksan21} use transformer networks to predict up to 20 seconds in the future, but only for cyclic motions. Zhou \emph{et al.}~\cite{Zhou18d} also target long term predictions, but provide only qualitative results for sequences from walking, dancing, and martial arts, which tend to follow well-structured patterns. Concurrent to our work Diller \emph{et al.}~ \cite{Diller22} use characteristic 3D poses resembling our keyposes for long-term motion prediction. However, these poses are manually annotated rather than automatically extracted from sequences. A different related task is to generate realistic motions by conditioning on the action label, rather than the past motion~\cite{Petrovich21,Chuan20} In our work, we show that regressing the future pose at every time instant is unnecessary and truly long-term prediction can be achieved more accurately by focusing the prediction on the essential poses, or keyposes, in a sequence. These poses are extracted automatically from the sequence, without manual annotations.

\vspace{-2.5mm}
\paragraph{Keyposes Applied to Other Tasks.}

Keyposes have been used for different tasks, such as action recognition. For example, in \cite{Lv07}, 2D keyposes are used for single view action recognition. In~\cite{Liu13b}, Adaboost is used to select keyposes that are discriminative for each action. In~\cite{Bloom17}, linear latent low-dimensional features extracted from sequences for action recognition and action prediction. Furthermore, \cite{Kovar02} focus on generating realistic transitions between nodes in a motion graph, which resembles our notion of keyposes, to synthesize short animated sequences. However, none of these works predict future keyposes given past ones.

\section{Methodology}

Classically, the task of motion prediction is defined as producing the sequence of 3D poses from $t=1$ to $t=N$, denoted as $\mathbf{P}_{1:N}$, given the sequence of poses from $t=-M$ to $t=0$, denoted as $\mathbf{P}_{-M:0}$.  Each pose value $\mathbf{P}_t$ is of dimension $3\times J$, where $J$ is the total number of joints. Therefore, motion prediction is written as 
\vspace{-1.5mm}
\begin{align*}
\mathbf{P}_{1:N} = F(\mathbf{P}_{-M:0})\;,
\end{align*}
where $F$ is the prediction function.

Our approach departs from this classical formalism by predicting keyposes from keyposes. As will be discussed in more detail in Section~\ref{sec:kp}, keyposes encode the important poses in a sequence $\mathbf{P}_{1:T}$, such that the remaining poses can be obtained by linear interpolation between subsequent keyposes. Therefore, our keypose-to-keypose framework takes as input a motion $\mathbf{P}_{-M:0}$ defined by its keyposes $\mathbf{K}_{-I_1:0}$, where $I_1 \ll M$ is the number of keyposes in the past sequence. We then predict $\mathbf{K}_{1:I_2}$, where $I_2 \ll N$ is the number of keyposes in the future sequence. We write this as
\vspace{-5mm}
\begin{align*}
\mathbf{K}_{1:I_2} = G(\mathbf{K}_{-I_1:0})\;,
\end{align*}
where $G$ is the keypose-to-keypose prediction function.

\begin{figure}[!] 
	\centering
	\vspace{-1mm}
	\includegraphics[width=1.0\linewidth]{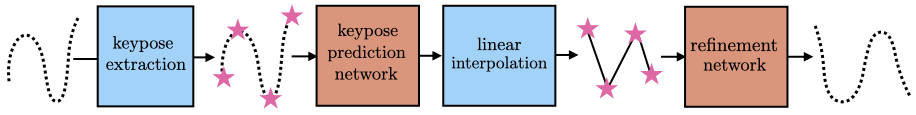}
	\vspace{-5mm}
	\label{fig:pipeline}
	\caption{ \textbf{Our overall pipeline} for predicting future motions via keyposes. It consists of the following steps: keypose extraction, keypose prediction, linear interpolation to reconstruct the sequence, and refining the final sequence.}
	\vspace{-4mm}
\end{figure}

Our overall pipeline, illustrated in Figure~\ref{fig:pipeline}, consists of extracting keyposes from input sequences, feeding them to the keypose prediction network, reconstructing the predicted sequence via linear interpolation, and refining the final result via a refinement network. We describe each of these steps in detail below.

\subsection{Keyposes}\label{sec:kp}
Let us now discuss how we obtain keyposes $\mathbf{K}_i$, $i \in [1, I]$, given a sequence of poses $\mathbf{P}_t$, $t \in [1, T]$. We define the keyposes as the poses in $\mathbf{P}_{1:T}$ between which linear interpolation can be used to obtain the remaining poses. We therefore employ an optimization-based strategy to identify the poses from which the L2 error between the original sequence $\mathbf{P}$ and the sequence reconstructed by linear interpolation is minimized.
Our method proceeds as follows:
\vspace{-2mm}
\begin{itemize}
	\itemsep-0.2em
	\item We set $\mathbf{P}_1$ and $\mathbf{P}_{T}$ to be the initial keyposes.
	\item We reconstruct the sequence by linearly interpolating the set of keyposes. We denote the reconstruction as $\mathbf{\hat{P}}_t$, $t \in [1,T]$. 
	\item We select the pose $\mathbf{P}_t$ at position $t$ which has the highest L2 error with respect to $\mathbf{\hat{P}}_t$, the pose reconstructed by linear interpolation at the same time index. We add  $\mathbf{P}_t$ to our set of keyposes.
	\item The algorithm continues recursively, selecting keyposes from the sequences between $[1,t]$ and $[t, T]$. The recursion ends once the average reconstruction error of the linear interpolation is below a threshold, yielding a set of keyposes.
\end{itemize}
\subsection{Motion Prediction with Keyposes}\label{sec:kp_clusters}
In principle, we could directly use the above-mentioned keyposes for prediction, by simply learning to regress keypose values. However, for long-term prediction, this would exhibit the same tendency as existing frameworks to converge to a static pose. To overcome this, we propose to cluster the training keyposes and treat keypose prediction as a classification task, where the clusters act as categories.

\begin{figure}[!]
	\centering
	\includegraphics[width=1.0\linewidth]{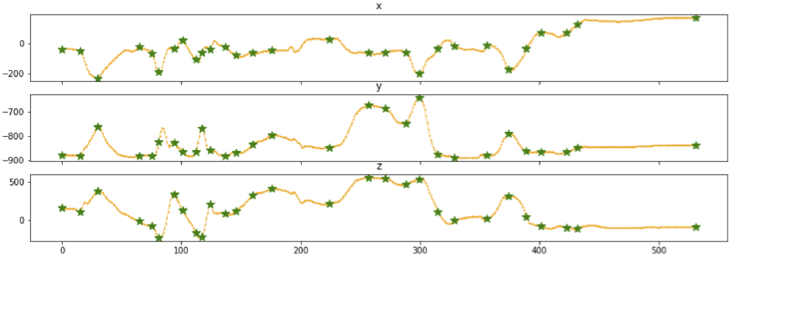}
	\vspace{-10mm}
	\caption{{\bf Distribution of keyposes} in a sequence from the Human3.6M dataset. The plots depict the x, y, and z coordinates of the right foot of Subject 1 during the purchases action. We show the locations of the keyposes as green stars, and the transition poses as orange dots. The downsampling effect is prominent. The keyposes are distributed more densely where the motion is the most varied, and these keyposes have shorter duration. Note that these plots only correspond to one joint, whereas the optimization takes into account the average error of all the joints.}
		\vspace{-4mm}
	\label{fig:keyposes}
\end{figure}

To this end, we extract the keyposes for every training motion individually, and cluster all the resulting training keyposes into $K$ clusters via k-means. Each keypose is then given a label determined by the cluster it is assigned to. Finally, we prune the keyposes by removing the unnecessary intermediate ones that have the same label as their preceding and succeeding keypose. 
An example distribution of keyposes in a sequence is shown in Figure~\ref{fig:keyposes}. 

This formalism allows us to cast keypose prediction as a classification problem. Specifically, instead of predicting the future keypose values, we predict their labels. Given the labels, $l_{i}$ and $l_{i+1}$, of two subsequent keyposes, $\mathbf{K}_{i}$ and $\mathbf{K}_{i+1}$, we can simply estimate the intermediate poses via linear interpolation between the corresponding cluster centers. However, this requires the duration $d_{i+1}$ between the two keyposes, indicating the number of intermediate poses, which we therefore also predict.

\subsubsection{Network Design and Training}

\begin{figure}[!]
	\centering
	\includegraphics[width=1.0\linewidth]{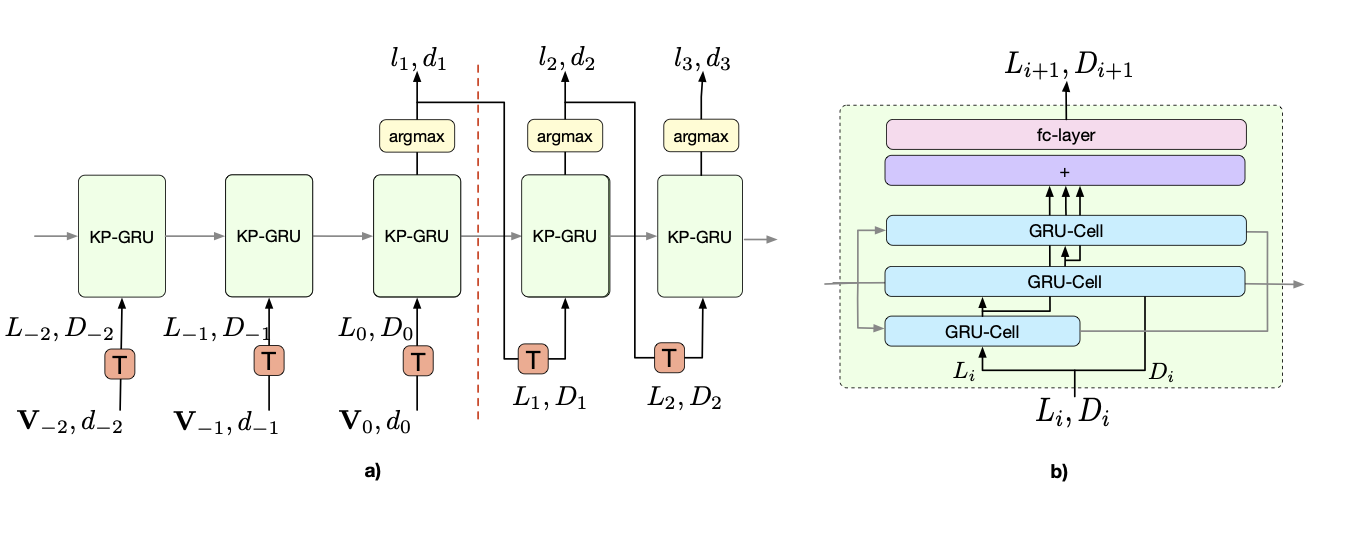}
	\vspace{-10mm}
	\caption{{\bf Keypose-to-keypose network structure}.(a) Overall architecture. At each time step $i$, a keypose GRU (KP-GRU) unit predicts the keypose labels and durations of the next step $i+1$. The time of the last observation is denoted by $i=0$. Before this time-step, the network is given ground-truth keyposes as conditioning signal. The label distribution $L_i$ for past keyposes is found using the keypose value $\mathbf{V}_i$. After time-step $i=0$, the network is given its own predictions as input rather than the ground truth. The label distribution $L_i$, in this case, is found using the predicted label $l_i$. The orange T blocks represent the transformation to compute the distributions. (b) Inner structure of the KP-GRU unit, which consists of a three layer GRU network followed by a fully connected layer.} 
		\vspace{-3mm}
	\label{fig:arch}
\end{figure}

We have designed an RNN based neural network as our keypose-to-keypose prediction framework, as shown in Figure~\ref{fig:arch}. 
At each time step, in addition to the hidden representation of the previous time step, our recurrent unit takes as input the previous keypose label $l_i$ and duration $d_i$. Specifically, we represent the label as a distribution $L_i$ computed as follows.
\vspace{-1mm}
\begin{enumerate}
	\itemsep-0.2em
	\item If we know the true keypose value (i.e., for observed past keyposes): We compute the proximity between the keypose value $\mathbf{V}_i$ and every cluster center $C_j$, $j \in [1,K]$ as the negative average Euclidean distance between the corresponding joints in $\mathbf{V}_i$ and $\mathbf{C}_j$. These values form a $K$-dimensional proximity vector for each keypose $i$.
	\item If we do not know the keypose value (i.e., for inferred future keyposes): We compute the proximities between the cluster center corresponding to the predicted label $l_i$, $\mathbf{C}_{l_i}$,
	and all cluster centers $C_j$, $j \in [1,K]$.
	\item We pass the resulting proximity vector through a softmax operation with a temperature of $0.03$ to obtain a distribution $L_i$ over the labels. 
\end{enumerate}
To also treat duration prediction as a classification task, we categorize the durations into very short (less than $4$ frames), short (between $5$ and $10$ frames), medium (between $10$ and $14$ frames), long (between $14$ and $25$ frames), and very long (more than $25$ frames). We then encode the duration $d_i$ of a keypose as a one-hot encoding $D_i$ over these categories and output a distribution for the future keyposes.

Therefore, our network predicts a pair of distributions: over the labels and over the duration categories. We train the network using two loss functions: 
\vspace{-0.5em}
\begin{itemize}
	\itemsep-0.2em
	\item $E_\text{labels}$: The cross-entropy loss between the ground-truth cluster label and the predicted label distribution;
	\item $E_\text{dur}$: The cross-entropy loss between the predicted duration distribution and ground-truth duration category. 
\end{itemize}
\vspace{-0.5em}

The overall loss of our network therefore is 
\vspace{-0.5em}
\begin{equation}
    E = w_\text{labels} E_\text{labels} +w_\text{dur} E_\text{dur}\;,
\end{equation}
where $w_\text{labels}$ and $w_\text{dur}$ weigh the different loss terms.

During training, the label of the next keypose $l_{i+1}$ is determined as the one with the highest predicted probability. We then compute a distribution $L_{i+1}$ from this label as described above. This procedure prevents error accumulation as the prediction progresses and guarantees that the network will never see anything very different from what it was trained on. The duration of the next keypose $d_{i+1}$ is determined similarly: According to the category with the highest probability, the duration is set to $3$ for very-short, $6$ for short, $12$ for medium, $16$ for long, and $25$ for very long. Using the predicted label and duration of each time-step, we can reconstruct the sequence via linear interpolation between the corresponding cluster centers, as described previously.

 During training, we observe $7$ past keyposes and predict $12$ future keyposes. At test time, we predict until we reach $5$ seconds. Weights of the loss terms are set to $w_\text{labels}=1.0$, $w_\text{dur}=0.1$. Our network is trained for $100$ epochs with a batch size of $64$. We use an Adam optimizer with a learning rate of $0.0001$ and a $0.01$ weight decay. We report the results of the model with the highest validation score. 

\subsubsection{Inference and Interpolation}

Our network produces distributions over the keypose clusters. Hence, at inference time, for each iteration of the recurrent network, we can sample the future label and duration from the predicted distributions. In practice, before sampling, we smooth the predicted distributions via a softmax with a temperature of $0.3$. This sampling scheme allows us to produce multiple future sequences given a single observation.

Once we have predicted a set of keypose labels and their durations, we can interpolate the intermediate poses and reconstruct the future sequence. Denoting by $t$ the time index of keypose $\mathbf{K}_{i}$ in the sequence, the intermediate pose at time $t_1>t$ is computed as
\vspace{-3mm}
\begin{align*}
\mathbf{P}_{t_1} =  \mathbf{C}_{l_{i}} + (t_1-t)\frac{\mathbf{C}_{l_{i+1}}-\mathbf{C}_{l_{i}}}{d_{i+1}}\;,
\end{align*}
where $\mathbf{C}_{l_{i}}$ and $\mathbf{C}_{l_{i+1}}$ are the cluster centers corresponding to labels $l_i$ and $l_{i+1}$.

The sequences obtained by linear interpolation can then be refined using a pretrained refinement network trained to produce sequences that preserve the poses of the original sequence. Formally this operation can be written as
	\vspace{-2mm}
	\begin{align*}
	\mathbf{P}_{1:N}^\text{ref} =  R({\mathbf{P}_{1:N}}) \; ,
	\end{align*}
where $R$ denotes the refinement function and $ \mathbf{P}_{1:N}^\text{ref}$ denotes the refined pose sequence. We describe this network in more detail in the appendix.

\section{Experiments}

\subsection{Datasets}

\textbf{Human3.6M}~\cite{Ionescu14a} is a standard 3D human pose dataset and has been widely used in the motion prediction literature~\cite{Martinez17b,Jain16,Mao19}. It contains $15$ actions performed by $7$ subjects. Human pose is represented using the 3D coordinates of $32$ joints. As previous work~\cite{Mao19,Lebailly20,Mao20}, we load the exponential map representation of the dataset, remove global rotation and translation, and generate the Cartesian 3D coordinates of each joint mapped onto a uniform skeleton. Following the implementation of existing works~\cite{Mao19,Lebailly20,Mao20}, subject $5$ is reserved for testing, subject $11$ for validation and the remaining subjects are used for training.  We test each method on the same $64$ sequences formed using indices randomly selected from Subject $5$'s sequences. Note that the observed keyposes are extracted using the sequence only up to the present time index as opposed to the entire sequence. The threshold used for keypose extraction is $500$mm, and we cluster the keyposes into $1000$ clusters.

\textbf{CMU-Mocap}~\cite{CMUHMC} is another standard benchmark dataset for motion prediction and was used in~\cite{Li20b,Mao19,Lebailly20}. As explained in~\cite{Li20b}, the eight action categories with enough trials are used for motion prediction. We used six out of eight actions, \emph{basketball, basketball signal, directing traffic, jumping, soccer, and wash window}, as the sequences for \emph{running and walking} were too short to provide enough input keyposes for our method. One sequence of each action is reserved for testing, one for validation and the rest are used for training. The dataset is loaded and processed in the same manner as Human3.6M. The threshold used for keypose extraction is $250$mm, as some sequences are quite short, and we found that extracting more keyposes increases validation accuracy. We cluster the keyposes into $100$ clusters, as this dataset is much smaller than Human3.6M and contains only $6$ action classes as opposed to $15$.

\subsection{Baselines} 

We selected the following baselines for comparison purposes:  HisRep~\cite{Mao20} and TIM-GCN~\cite{Lebailly20} constitute the SOTA among the methods designed for long-term prediction. For HisRep, we evaluate two versions. The first one, HisRep10, was presented as the best model in~\cite{Mao20}. It is trained to output $10$ frames and iteratively use the predicted frames as input for longer term prediction. We also evaluate HisRep125, which directly predicts $125$ frames by taking $150$ past frames as input. For TIM-GCN, we trained a model that observes subsequences of lengths $10$, $50$ and $100$ and predicts $125$ frames, hence tailoring the architecture to longer-term predictions of $5$ seconds. Finally we compare against Mix\&Match~\cite{Aliakbarian20} and DLow~\cite{Yuan20}, the SOTA methods for multiple long-term motion prediction, trained to predict $125$ future frames using $100$ past frames. For all the baselines, we used the model that gave the best validation accuracy, to be consistent with our model selection strategy.

\subsection{Metrics}

As in~\cite{Ghosh17}, we evaluate the quality and plausibility of the generated motions by passing them through an action classifier trained to predict the action category of a given motion. If the predicted motion is plausible, such a classifier should output the correct class. To focus our evaluation on the quality of the predicted {\it motions},  we designed a Motion-Only Action Classifier (MOAC) based on the architecture of~\cite{Li19m}, with the pose stream removed and only the motion stream remaining. It takes as input motions encoded as the difference between poses in consecutive time-steps. This eliminates the scenario of a static prediction scoring very high under this metric. We have trained it on the training sequences of Human3.6M and CMU-Mocap separately. We report the top-K action recognition accuracy in percentages obtained with this classifier. For our method, Mix\&Match, and DLow, which can output multiple future predictions, we report the average accuracy over $100$ predictions. 
	
We also report the PSKL metric~\cite{Ruiz19}, which is the KL divergence between the power spectrums of the ground-truth future motions and the predictions. As the KL divergence is asymmetric, we evaluate it in both directions and denote the results as `gt-pred' and `pred-gt' respectively. These values being close indicates that the ground truth and predicted motions are similarly complex.

The mean per-joint position error (MPJPE) is the most commonly used metric to evaluate motion prediction. We report the MPJPE errors at 1 second, which is the conventional long-term timestamp, and at 5 seconds. For multiple-prediction methods, we report the MPJPE results of the closest predicted sequences. We present two results: the MPJPE calculated by finding the sequence with the minimum \textit{average} MPJPE (denoted as ``ave") and the sequence with the minimum MPJPE at the second being evaluated (denoted as ``best").

Finally, for multiple-prediction methods we report the results of a diversity metric~\cite{Yuan20,Aliakbarian20} for 100 predictions, calculated by finding the average pairwise $L2$ distances between all pairs of generated sequences.

\subsection{Comparative Results}

\begin{figure*}[!]
	\centering
	\includegraphics[width=1.0\linewidth]{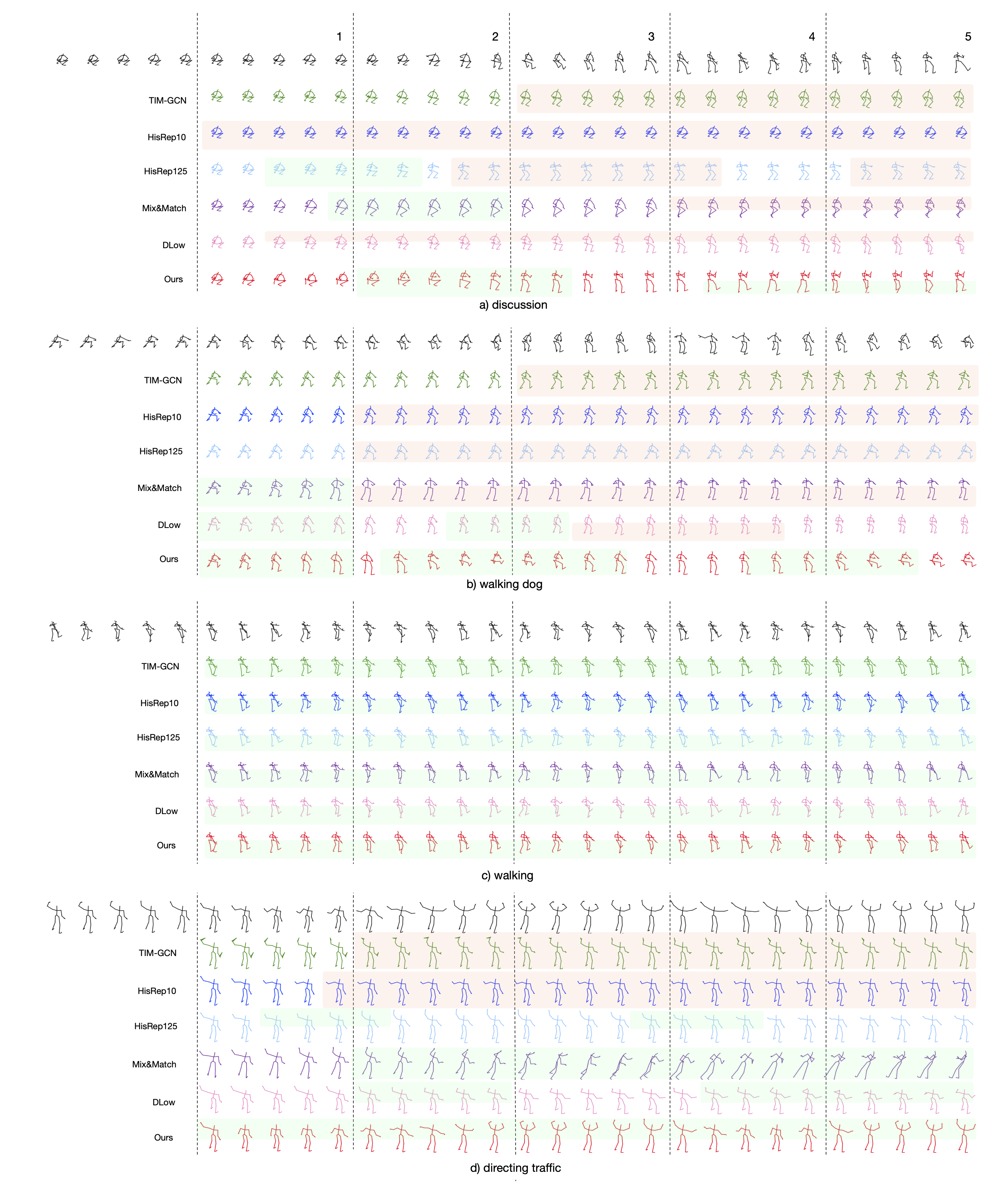}
	\vspace{-10mm}
	\caption{{\bf Qualitative evaluation of our results }on the Human3.6M (Figures a, b, c) and CMU-Mocap (Figure d) datasets. We present the results of: TIM-GCN (green), HisRep10 (dark blue), HisRep125 (light blue), Mix\&Match (violet), DLow (pink), Ours (red). For the multiple prediction methods, we display the prediction that has the lowest average MPJPE error with respect to the ground truth. The top black row depicts the ground truth, and the first $5$ poses are the conditioning ones. The numbers at the top indicate the future timestamp in seconds. We highlight the segments and body parts that undergo significant motion in green, and the areas that are static for long stretches in red. Our approach yields more dynamic poses for discussion, walking dog and directing traffic, which are acyclic motions. For cyclic motions such as walking, the other methods are also able to produce dynamic poses.}
	\label{fig:qualitative}
\end{figure*}

\begin{figure*}[!]
	\centering
	
			\vspace{-6mm}
	\includegraphics[width=1.0\linewidth]{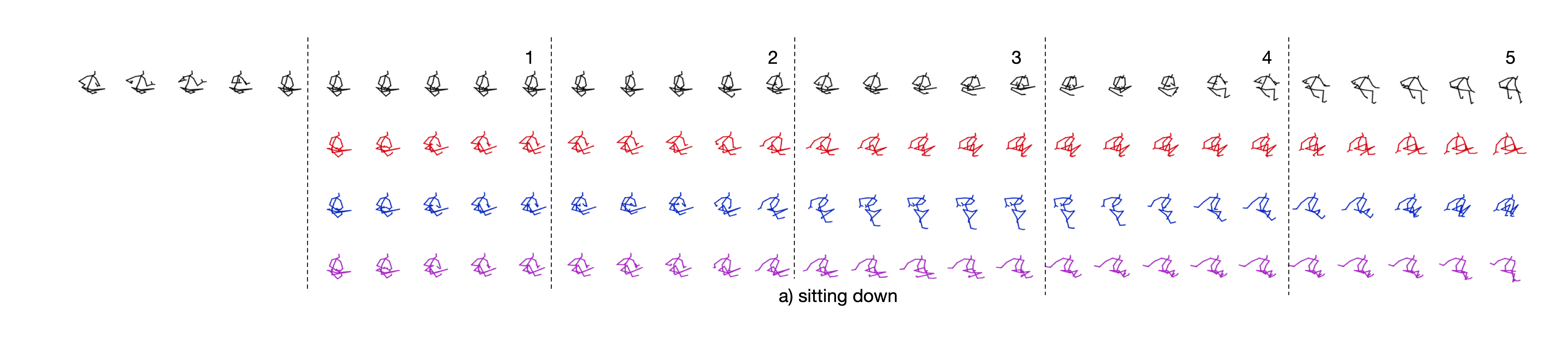}
		\vspace{-9mm}
	\caption{{\bf Qualitative results of our multiple motion prediction} obtained by sampling the predicted label distribution. The numbers at the top indicate the future timestamp in seconds. The top row in black depicts the ground truth, and the remaining rows in color are our multiple generated motions. The sampled motions are diverse, yet can all still be classified as ``sitting down". }
	\vspace{-1mm}
	\label{fig:diverse} 
\end{figure*}

\begin{table*}[t]
	\centering
	\small{
\begin{tabular}{|l|c|c|c|c|c|c|c|c|}
	\hline
	& \multicolumn{4}{c|}{Human3.6M} & \multicolumn{4}{c|}{CMU-MoCap} \\
	\hline                         &top-1 & top-2 & top-3 & top-5&top-1 & top-2 & top-3 & top-5   \\ \hline
	
	oracle							 &  51  &  70    & 79  &91&86&88&90&100  \\  \hline
	TIM-GCN~\cite{Lebailly20}              & 16   &  26    &  36   & 55 &44&69&85&95 \\ \hline
	HisRep10~\cite{Mao20}                     & 21 & 32  &  39 & 53 &42&54&62&88  \\ \hline
	HisRep125~\cite{Mao20}                    & 20& 32 & 44 &60  & 34&48&57&82\\ \hline
	Mix\&Match~\cite{Aliakbarian20}                &  18  & 32     & 45  & 61 &30&39&58&85  \\ \hline
	 DLow~\cite{Yuan20}   &  16 & 26     & 39   & 56  &36&49&60&79\\ \hline
	 Ours   & \textbf{32}& \textbf{44}& \textbf{54} & \textbf{69} & \textbf{74}& \textbf{81}& \textbf{88} & \textbf{99}  \\ \hline
	 
	\end{tabular}}

\caption{{\bf Results of the motion-only action classifier (MOAC) on the Human3.6M and CMU-MoCap datasets}. We compare the classification accuracies for the motions predicted with our method and with the SOTA ones. We also report the accuracies of the oracle, which evaluates the ground-truth future motions, as an upper bound. We report the top-1, top-2, top-3 and top-5 accuracies. The results indicate that the motions predicted by our keypose network are more realistic than those produced by the competing methods.}
	\vspace{-3mm}
\label{tab:moac}

\end{table*}

\begin{table*}[!h]
	\centering
	\small{
		\begin{tabular}{|l|c|c|c|c|c|c|c|c|}
			\hline
			& \multicolumn{4}{c|}{Human3.6M} & \multicolumn{4}{c|}{CMU-MoCap} \\
			\hline                         &gt-pred& pred-gt &average&difference &gt-pred& pred-gt&average&difference  \\ \hline
			
			TIM-GCN~\cite{Lebailly20}                   &  0.0069 & 0.0098 &0.0083 & 0.0029 &  0.0073 & 	0.0101   &0.0087&0.0028 \\ \hline
			HisRep10~\cite{Mao20}                      & 0.0076  & 0.0129 &0.0103&0.0053  & 0.0061 & 0.0081  &0.0071&0.0020 \\ \hline 
			HisRep125~\cite{Mao20}                    & 0.0070  & 0.0097  &0.0083&0.0027  & 0.0065 & 0.0093 &0.0079&0.0028  \\ \hline
			Mix\&Match~\cite{Aliakbarian20}          &  0.0067 & 0.0075  &0.0071& 0.0008 & 0.0090 & 0.0104  &0.0097&0.0014   \\ \hline
			DLow~\cite{Yuan20}           & 0.0062 &0.0080  &0.0071& 0.0018& 0.0069 & 0.0073   &0.0071&0.0008   \\ \hline
			Ours                       & \textbf{0.0059}   & \textbf{0.0061}  &\textbf{0.0060}& \textbf{0.0002} &\textbf{0.0057} & \textbf{0.0062} &\textbf{0.0059} & \textbf{0.0005} \\ \hline
	\end{tabular}}
	\caption{ {\bf PSKL results} on both the Human3.6M and CMU-MoCap datasets; lower numbers indicate better results. We report the PSKL values between ground truth and predictions (`gt-pred') and vice-versa (`pred-gt'),  their average, and their absolute difference. For the multiple prediction methods, Mix\&Match, DLow and Ours, we report the best PSKL value, obtained from the predictions that have the most similar power spectrum to the ground truth future motion. We observe that the trend is similar to the MOAC results, with our method outperforming the  SOTA.}
	\label{tab:PSKL}
	\vspace{-3mm}
\end{table*}

\begin{table*}[!h]
	\centering
	\small{
	
		\begin{tabular}{|l|c|c|c|c|c|c|}
			\hline                         &diversity $\uparrow$&  accuracy $\uparrow$&1s ave $\downarrow$&1s best $\downarrow$&5s ave $\downarrow$&5s best $\downarrow$
			\\ \hline
			TIM-GCN~\cite{Lebailly20}       & - &16  &143&143 &196&196  \\ \hline
			HisRep10~\cite{Mao20}          & - &21  &\textbf{116}&\textbf{116}&197&197  \\ \hline
			HisRep125~\cite{Mao20}          & - &20  &136&136&191&191  \\ \hline
			Mix\&Match~\cite{Aliakbarian20}          & 1002 &18  &161&156 &244&237  \\ \hline
			DLow~\cite{Yuan20}          & 3501 &16  &136& 131 &\textbf{189}& 171 \\ \hline
			Ours (0.1)                    & 6936 &\textbf{34}  &177&168 &208&173 \\ \hline
			\bf{Ours (0.3)}                    & 10328 & 32   &157&138&196&151 \\ \hline
			Ours (0.5)                   & 12362 &30  &154&125 &191&137 \\ \hline
			Ours (0.7)                    & 13491 &27  &144&118 &\textbf{190}&130 \\ \hline
			Ours (1.0)                  & \textbf{14995} &20&145&\textbf{116} &194&\textbf{127} \\ \hline
	\end{tabular}
	\caption{ {\bf Results on the diversity metric, top-1 MOAC accuracy and MPJPE errors} on the Human3.6 dataset. Higher diversity values indicate more variation in the multiple future predictions and lower MPJPE values indicate closer predictions to ground truth future motion. We have highlighted close best results in bold. We provide several results of our method with varying sampling softmax temperature, indicated in parentheses. As the temperature increases, the diversity values of the predictions increase and MPJPE values decrease, however the average top-1 MOAC accuracy begins to decrease as well.}
	\label{tab:diverse}
	\vspace{-5mm}
}
\end{table*}

We compare our approach to the baselines on Human3.6M and CMU-Mocap in Table~\ref{tab:moac} on the MOAC metric. In both cases, our method outperforms the others by a large margin. In Table~\ref{tab:PSKL}, we report the results for the PSKL metric and show that we outperform the other methods by having both lower PSKL values and having very close `gt-pred` and `pred-gt` values.

In Table~\ref{tab:diverse}, we evaluate the diversity and MPJPE losses of the predicted sequences. We observe that the diversity value of our method increases as we increase the softmax temperature used for sampling during inference. Increased diversity allows us to achieve lower MPJPE values since we now have a higher chance of sampling the correct future motion. However this also leads to a drop in average MOAC accuracy. This clearly shows the tradeoff between predicting diverse motions and motions that represent the action of interest, or are close to the ground truth. Therefore, in our evaluations we choose to set the temparature to $0.3$, trading a bit of accuracy for more diverse predictions. Our method outperforms the others in having both high diversity, the best average MOAC accuracies, and low MPJPE. For MPJPE, at 1 second we are comparable to the other methods, but at 5 seconds, especially for the ``best" MPJPE, our performance is noticeably better. 

Note that we present the MPJPE results to give a complete picture but do not believe it to be the best metric for evaluating long term prediction methods, especially for acyclic motions. Consider, for example, the walking dog action, where the subject, in the middle of the walk, kneels down to pet the dog and stands back up. Our method, in contrast to others, is able to predict the order of these motions, as reflected by our high MOAC score. By contrast, the MPJPE is highly sensitive to the timing of the motions and can be thrown off by slight shifts in timing. For instance, the MPJPE error between two phase shifted sinusoidals, or sinusoidals of slightly different frequencies, would be high. For such cases, the MPJPE between a flat signal and a sinusoidal might even be lower, but the flat signal would be completely incorrect.

Fig.~\ref{fig:qualitative} depicts qualitative results for the discussion, walking dog and walking actions of Human3.6M, and for the directing traffic action of CMU-Mocap. Close visual inspection reveals that, while all methods work reasonably well on cyclic motions such as walking, ours does better on the acyclic ones, such as walking dog. It produces wider motion ranges than the others that tend to predict less dynamic motions. Fig.~\ref{fig:diverse} depicts qualitative results for multiple predictions. Our method is capable of generating diverse, yet still plausible motions.

\subsection{Ablation Study on Keypose Retrieval Methods}
We evaluate the effect of using keyposes obtained via different strategies: sampling, clustering and ours. The naive-sampling method evenly samples the motion every $15$ frames, which is the average keypose duration from our method. The keyposes are then clustered without any keypose pruning. This method also doesn't require predicting durations, as the duration will always be $15$. We also evaluate naive-sampling-pruned, where the keyposes are found through naive sampling, and then pruned. The clustering method performs clustering on \textit{every} pose in the sequence, rather than only on the poses found via our optimization strategy and pruned afterwards.

As shown in Table~\ref{tab:ablation_kp}, our keypose method achieves the highest MOAC accuracies. The comparison with the naive-sampling method emphasizes the importance of having variable-duration keyposes, as opposed to evenly sampling the motion. The comparison with the clustering method emphasizes the importance of optimizing for the keyposes.

\begin{table}[t]

	\centering
	\small{

		\begin{tabular}{|l|c|c|c|c|c|}
	\hline		&top-1 & top-2 & top-3  & top-5 \\ \hline

			Naive-sampling       &  28   &   38    &    51& 67 \\ \hline
			Naive-sampling-pruned &  30   &   42    &    52& 63\\ \hline
			Clustering           &  24 &   37 & 48   &  66\\ \hline
			Ours        & \textbf{32}& \textbf{44}& \textbf{54} & \textbf{69} \\ \hline
			
	\end{tabular}}

	\caption{{\bf Analysis of the method to obtain keyposes.} We compare the MOAC accuracies of different keypose methods.
	Our method achieves higher classification accuracies than the other ones, indicating that the quality of the keyposes affects the performance.}
	\label{tab:ablation_kp}
	\vspace{-2mm}
\end{table}

\subsection{Limitations and Failure Modes}
\label{sec:failure}

\begin{figure}[t]

	\centering
	\includegraphics[width=1.0\linewidth]{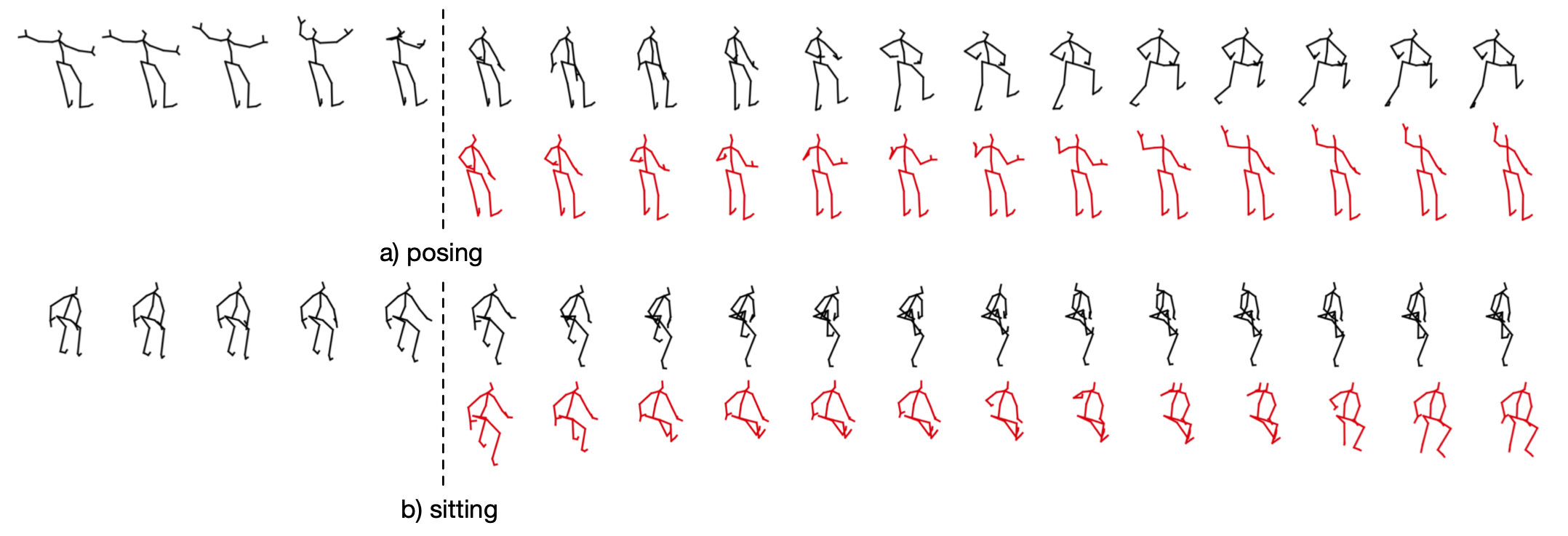}
	\vspace{-4mm}
	\caption{{\bf Examples of failure} to predict the correct keypose labels. Top: Example from the ``posing" category. Once our model detects the extended arms, it switches to the waving motion resembling the poses from the ``greeting" category. Bottom: Example from the ``sitting" category. Although the leg motion is plausible, our prediction lifts a hand to its head, resembling a motion from the ``phoning" category.}
	\vspace{-4mm}
	\label{fig:failure}
\end{figure}

The main failure mode of our approach arises from incorrect cluster label prediction, as illustrated in Figure~\ref{fig:failure}, and from the fact that, while powerful, cluster centers cannot perfectly model all poses. To overcome this, we will study in future work the use of other clustering strategies such as the deep-learning based one of~\cite{Oord17} that can be incorporated into our architecture for end-to-end training.

\section{Conclusion}

We have presented an approach to long-term motion prediction. To the best of our knowledge, our work constitutes the first attempt at long-term prediction out to $5$ seconds in the future. To this end, we have reformulated motion prediction as a classification problem that guesses in which one of a set of keypose clusters the subject will be. To validate our approach, we have introduced a new action classifier, MOAC, that specifically focuses on the transitions between poses, thus placing an emphasis on the correctness of \textit{motion}, rather than that of \textit{poses}. Our experiments show that our method yields more dynamic and realistic poses than state-of-the-art techniques, even when they are tailored to learn patterns for long-term prediction. Furthermore, our approach lets us easily propose multiple possible outcomes.

Altogether, we believe that our approach could be highly beneficial for autonomous systems, such as an autonomous car that needs more than $1$ second window into the future to react to pedestrian motions. Furthermore, the ability to sample many alternative future situations can be exploited to aid the motion planning of autonomous systems. Ultimately, long-term and short-term predictions should be used in a complementary manner, the former to produce long-term probabilistic scenarios for better action planning, and the latter to predict fine details in the immediate future.

\textbf{Acknowledgements.} This work was supported in part by the Swiss National Science Foundation.

{\small
	\bibliographystyle{ieee_fullname}
	\bibliography{string,vision,graphics,learning,robotics,misc}
}

\section{Appendix}

\subsection{Video}
The video, which can be accessed on our project page \url{https://senakicir.github.io/projects/keyposes} includes a short overview of our work, summarizing our motivation, methodology and results. 

\subsection{Ablation Studies}

In our ablation studies, we report the MOAC accuracy results on the validation set as well as the test set. The validation accuracy reported is the sum of top-1, top-2, top-3 and top-5 OMAC accuracies on the validation set. For all cases, we chose our final model as the one obtaining the highest validation accuracy. Please note that the results presented here are the not the average accuracies of multiple predicted motions. In order to evaluate quickly, we chose to predict a single future in a ``greedy" manner: instead of sampling multiple futures from the predicted cluster probability distribution, we simply choose the one with the highest probability. We then report the average accuracies of these single predictions.

\paragraph{Number of clusters} We report results of using different number of clusters in Table~\ref{tab:ablation_cluster}. Using too few clusters leads to both lower validation and test accuracies. Using more clusters doesn't lead to a major drop in accuracies but leads to an increase in training time. We chose to use $1000$ clusters in our final model.

\begin{table*}[!h]
\centering{
\begin{tabular}{|l|c|c|c|c|c|c|}
\hline                     &top-1 & top-2 & top-3 & top-5&val-acc &training-time (hours)   \\ \hline
100                      & 25& 37  & 49   &63  &234&1.4 \\ \hline
250                      & 24& 38 & 53& 67    & 252&1.5\\ \hline 
500                     &31&43&	53&\textbf{68} &251&1.8\\ \hline
1000 (\textbf{Ours})  &\textbf{34}& \textbf{46}  &\textbf{55}   &68  & \textbf{264} &2.0\\ \hline
1500                      &\textbf{34}&44&\textbf{55} &\textbf{70} &	261&2.1\\ \hline
2000                      &\textbf{34}&43&\textbf{55} &\textbf{70} &	263&2.2\\ \hline

\end{tabular}}

\caption{{\bf Ablation study on number of cluster centers} We observe that we have similar performance around using $1000$ clusters, with $1000$ giving the highest accuracy. As the number of clusters decreases, the training-time also decreases but the accuracy drops. This is caused by the final layer of the keypose network being a fully connected one, going from the hidden layer size ($512$) to the number of clusters. A decrease in the number of clusters therefore leads to a smaller network which is faster to train. An increased number of clusters also results in similar accuracies but an increase in training-time.}
\label{tab:ablation_cluster}
\end{table*}

\paragraph{Keypose Error Threshold} The keypose extraction algorithm recursively selects keyposes from the sequence until the reconstruction error is below a threshold. In our experiments, we chose this number to be $500$. We report results of using different error thresholds in extracting keyposes in Table~\ref{tab:ablation_kp_threshold}. Using a threshold that is too low or high leads to lower validation accuracies. We also note that using a higher threshold leads to sparser keyposes, leading to a smaller training set. This allows for faster training, however the accuracy drops significantly when this threshold is set too high.

\begin{table*}[htbp]
\centering{
\begin{tabular}{|l|c|c|c|c|c|c|c|}
\hline                     &top-1 & top-2 & top-3 & top-5&val-acc&training set size&training-time (hours)   \\ \hline
250                      &29	&38&	49&	66&	263&13823&2.6\\ \hline
500 (\textbf{Ours})                     & \textbf{34}& \textbf{46}  &\textbf{55}   & \textbf{68}  & \textbf{264} &8894&2.0\\ \hline
750                      &32&	42&	53	&63	&262&5417&1.4\\ \hline
1000                     &31&42&50&64&242&3036&1.2 \\ \hline
1500                     &12&22&31&46&169&396&0.5 \\ \hline
\end{tabular}}

\caption{{\bf Ablation study on the error threshold for the keypose extraction algorithm.} We observe that we have similar performance around using $500$ as the error threshold, with $500$ giving the highest accuracy. As the error threshold decreases, the training-time increases because the size of the training set increases. As the error threshold increases, the training-time decreases as the set of keyposes becomes sparser. However the validation accuracy drops dramatically.}
\label{tab:ablation_kp_threshold}
\end{table*}

\paragraph{Noise} During training, we add noise to the proximity vectors as a part of calculating the label distribution. In Table \ref{tab:noise_magnitude} we analyze the effects of this noise. We find that, indeed, adding a noise with standard deviation $0.1$ is useful and increases accuracies.

\begin{table*}[!h]
	\centering{
		\begin{tabular}{|l|c|c|c|c|c|}
			\hline                     &top-1 & top-2 & top-3 & top-5&val-acc    \\ \hline
			no-noise                      & 24& 35  &45& 62   & 239\\ \hline
			noise $=0.05$                     &30 &42   & 53   & 679& 261\\ \hline 
			noise $=0.1$ \textbf{(Ours)}    & \textbf{34}& \textbf{46}  &\textbf{55}   & 68 & \textbf{264} \\ \hline
			noise $=0.5$                  & 19 & 29  & 39   &55  &172\\ \hline   
			noise $=1.0$                &13& 23& 33&50&146\\ \hline
	\end{tabular}}
	
	\caption{{\bf Ablation study on the standard deviation of the Gaussian noise} added to the proximities during keypose tokenization. We observe that adding a noise of $1$ increases accuracies, while using too much or too little noise leads to worse results.}
	\label{tab:noise_magnitude}
\end{table*}

\paragraph{Hidden Layer Size} Table \ref{tab:hidden_size} reports the results of using different hidden layer sizes. We choose $512$ as it leads to higher accuracies.

\begin{table*}[!h]
	\centering{
		\begin{tabular}{|l|c|c|c|c|c|}
			\hline                         &top-1 & top-2 & top-3 & top-5&val-acc    \\ \hline
			hidden size $=256$       & 29& 40  & 51   &65 & 236\\ \hline
			hidden size $=512$ \textbf{(Ours)}      & \textbf{34}& \textbf{46}  &\textbf{55}   & 68 & \textbf{264}  \\ \hline
			hidden size $=1024$     & 32& 42  & 54   & \textbf{71} & 261\\ \hline 
	\end{tabular}}
	
	\caption{{\bf Ablation study on the hidden layer size of the GRU cells.} We observe that using $512$ as the hidden layer size leads to higher accuracies.}
	\label{tab:hidden_size}
\end{table*}

\paragraph{Scheduled Sampling} For scheduled sampling, we use inverse sigmoid decay,
\begin{equation}
\epsilon_i = k/(k + \exp(i/k))
\end{equation}
as presented in~\cite{Bengio15b} where $k$ is the hyper-parameter of scheduled sampling and $\epsilon_i$ is the probability that teacher forcing is performed in that iteration. Table~\ref{tab:scheduled_sampling} reports the results of using scheduled sampling for teacher forcing with varying $k$. We conclude that using scheduled sampling with $k=10$ gives the highest accuracies.

\begin{table*}[!h]
\centering{
\begin{tabular}{|l|c|c|c|c|c|}
\hline                     &top-1 & top-2 & top-3 & top-5&val-acc    \\ \hline
No-tf                    &32 & 42  & 51   & 67 &\textbf{264} \\ \hline 
Always-tf 					  & 31& 42 & 53  & \textbf{70}  &250\\ \hline  
$k=10$ \textbf{(Ours)}   & \textbf{34}& \textbf{46}  &\textbf{55}   & 68 & \textbf{264} \\ \hline
$k=30$                     &30&41&54& \textbf{70} &259 \\ \hline 
$k=60$                & 32& 41 &51   &67  &260\\ \hline
$k=100 $                     & 29& 41  & 50   & 66 &259\\ \hline 

\end{tabular}}

\caption{{\bf Ablation study on the scheduled sampling} used for determining the probability of teacher forcing. We first evaluate the cases of not using scheduled sampling at all: either by not using teacher forcing at all during training (first row), or always using teacher forcing during training (second row). We then evaluate different cases of using scheduled sampling by varying the scheduled-sampling hyper-parameter $k$. We observe that using scheduled sampling and setting the parameter $k$ to $10$ and not using teacher forcing at all gives the same best results in terms of validation accuracy. In our experiments we chose to use scheduled sampling with $k=10$.}
\label{tab:scheduled_sampling}
\end{table*}

\paragraph{Training Softmax Heat} During training, we tokenize our keyposes by passing the inverse distances of the keypose value to the cluster centers through a heated softmax function. In Table~\ref{tab:softmax_heat}, we conduct an ablation study on the value of the heat. We observe that too much heat and too little heat both lead to low validation and test accuracies. In our experiments we set the softmax heat for training to $0.03.$

\begin{table*}[!h]
	\centering{
		\begin{tabular}{|l|c|c|c|c|c|}
			\hline                     &top-1 & top-2 & top-3 & top-5&val-acc    \\ \hline
			heat $=0.01$                     & 30 & 43  & 54  & \textbf{72} & 258\\ \hline
			heat $=0.03$ \textbf{(Ours)}     & \textbf{34}& \textbf{46}  &\textbf{55}   & 68 & \textbf{264}\\ \hline
			heat $=0.1$                      & 26&   37&  49 & 66& 238  \\ \hline 
			heat $=1.0$                      & 12& 24  & 35 & 53 &139\\\hline
	\end{tabular}}
	
	\caption{{\bf Ablation study on the heat used during the heated softmax} operation of the label tokenization. We use the heat $03$ during training as it gives the highest validation accuracy, which also corresponds to the high test accuracies. We observe that when the heat is too large, the validation accuracy drops significantly.}
	\label{tab:softmax_heat}
\end{table*}

\paragraph{Past Supervision} In Table~\ref{tab:past_supervision}, we analyze whether enforcing the loss on the outputs of the network as it processes the conditioning ground truth increases accuracy. We observe that this training strategy slightly increases both validation and test accuracies. 

\begin{table*}[!h]
	\centering{
		\begin{tabular}{|l|c|c|c|c|c|}
			\hline                     &top-1 & top-2 & top-3 & top-5&val-acc    \\ \hline
			No-past-supervision                      & 33& 43  & 54   &67  &263 \\ \hline
			With-past-supervision  \textbf{(Ours)}      & \textbf{34}& \textbf{46}  &\textbf{55}   & \textbf{68} & \textbf{264}  \\ \hline
	\end{tabular}}
	
	\caption{{\bf Ablation study on supervising already seen keyposes}. We observe that supervising on the conditioning past slightly improves validation and test accuracies.}
	\label{tab:past_supervision}
\end{table*}

\paragraph{Weights of loss functions} Table \ref{tab:classification_losses} reports the results of using different weights for the two loss functions we use. We find that setting these weights to $w_\text{ce}=1.0$ and $w_\text{dur}=0.1$ leads to the highest accuracies.

\begin{table*}[!h]
	\centering{
		\begin{tabular}{|l|c|c|c|c|c|}
			\hline                     &top-1 & top-2 & top-3 & top-5&val-acc    \\ \hline
			$w_\text{ce}=0.1, w_\text{dur}=1.0$                     & 28& 39  &48& 65   & 220\\ \hline
		   $w_\text{ce}=1.0, w_\text{dur}=0.1$  \textbf{(Ours)}          & \textbf{34}& \textbf{46}  &\textbf{55}   & \textbf{68} & \textbf{264} \\ \hline 
			$w_\text{ce}=1.0, w_\text{dur}=1.0$         & \textbf{34}& 44  &54& \textbf{68}   & 260\\ \hline
	\end{tabular}}
	
	\caption{{\bf Ablation study on the weights of different loss functions we use during training.} In the end, we chose the weight proportions of $w_\text{ce}=1.0$,$w_\text{dur}=0.1$ as this gave the highest validation accuracy.}

	\label{tab:classification_losses}
\end{table*}

\subsection{Action Specific Results}
We provide the top-3 MOAC results on each action in Human3.6M in Table~\ref{tab:action_rebuttal} in terms of top-3 MOAC results. We compare to HisRep10~\cite{Mao20}, as it proved to be the best performing SOTA on the MOAC metric. For $12$ out of $15$ classes, we outperform HisRep10. We note that on ``walking", which is cyclical, both methods perform equally well, as hinted at by the presented qualitative results.

\begin{table}[h!]
\centering
\footnotesize

\begin{tabular}{|l|l|l|l|l|}
	\hline
	\rowcolor[HTML]{C0C0C0} 
	& Walking      & Eating      & Smoking                                                & Discussion                                             \\ \hline
	HisRep10 & 97           & 30          & 28                                                     & 33                                                     \\ \hline
	Ours     & \textbf{100} & \textbf{83} & \textbf{47}                                            & \textbf{55}                                            \\ \hline
	\rowcolor[HTML]{C0C0C0} 
	& Directions   & Greeting    & Phoning                                                & Posing                                                 \\ \hline
	HisRep10 & 0            & 0           & 52                                                     & 23                                                     \\ \hline
	Ours     & \textbf{80}  & \textbf{27} & \textbf{63}                                            & \textbf{45}                                            \\ \hline
	\rowcolor[HTML]{C0C0C0} 
	& Purchases    & Sitting     & \begin{tabular}[c]{@{}l@{}}Sitting\\ Down\end{tabular} & \begin{tabular}[c]{@{}l@{}}Taking\\ Photo\end{tabular} \\ \hline
	HisRep10 & 5            & \textbf{98} & 56                                                     & 13                                                     \\ \hline
	Ours     & \textbf{33}  & 42          & \textbf{80}                                            & \textbf{23}                                            \\ \hline
	\rowcolor[HTML]{C0C0C0} 
	& Waiting      & W. Dog      & W. Together                                            &                                                        \\ \hline
	HisRep10 & \textbf{53}  & 6           & \textbf{94}                                            &                                                        \\ \hline
	Ours     & 34           & \textbf{86} & 36                                                     &                                                        \\ \hline
\end{tabular}

\caption{We compare the top-3 MOAC accuracies to HisRep10~\cite{Mao20}, the baseline that delivered the strongest MOAC results on Human3.6M. We outperform the SOTA method in $12$ of the $15$ action categories.}
\label{tab:action_rebuttal}

\end{table}

\subsection{Refinement Network} 
The refinement network is used to improve the results qualitatively. It takes as input the  sequences obtained by linear interpolation of the keypose cluster centers and their durations. We recall that the linear interpolation operation to find the pose $\mathbf{P}_{t_1}$ can be expressed as
\begin{align}
\mathbf{P}_{t_1} =  \mathbf{C}_{l_{i}} + (t_1-t)\frac{\mathbf{C}_{l_{i+1}}-\mathbf{C}_{l_{i}}}{d_{i+1}}\;,
\end{align}%
where $\mathbf{C}_{l_{i}}$ and $\mathbf{C}_{l_{i+1}}$ are the cluster centers corresponding to the predicted labels $l_i$ and $l_{i+1}$. $d_{i_+1}$ corresponds to the duration between the two keyposes at $i$ and $i+1$. We use this operation to reconstruct the entire predicted sequence 	$\mathbf{P}_{1:N}$.

The refinement network is a pretrained network which takes $\mathbf{P}_{1:N}$ as input and refines this result during inference. The resulting sequences are qualitatively much improved: they appear a lot more natural as they are smoother and contain less abrupt motions. We present results in the supplementary video. We have also compared the MOAC accuracies of the predicted sequences compiled using only linear interpolation versus additionally using the refinement network and found them to be very similar. The quantitative results reported in the main paper are obtained using linear interpolation.

\begin{figure}[!h]
	\centering
	\includegraphics[width=1.0\linewidth]{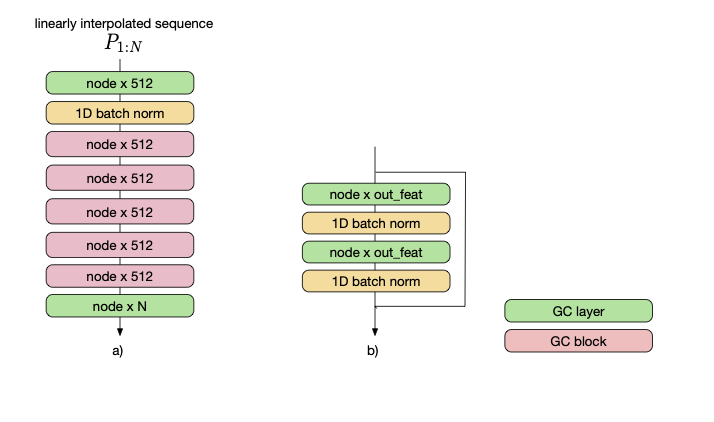}
	\caption{{\bf Refinement network model} based on~\cite{Mao19}. a) The architecture of the refinement network, where the green blocks represent graph convolution layers (GC layer) and pink blocks represent graph convolution blocks (GC block). b) The contents of a GC block. The dimensions written inside the blocks are the nodes and the number of output features, respectively, of each GC layer and GC block. We have $66$ nodes for H36m and $96$ nodes for CMU-MoCap datasets, equal to the number of joint trajectories for each dataset. The final number of output features $N$ is set to $125$, the number of frames in a $5$ second sequence.}
	\label{fig:refinement_arch}
\end{figure}

The architecture of the refinement network is similar to that of the graph convolutional network (GCN) architecture proposed by Mao \textit{et al.}~\cite{Mao19} for short term human motion prediction. We have found that a GCN architecture fits well for capturing the relationship between joint trajectories for the refinement of fine-details. Figure~\ref{fig:refinement_arch} depicts our network architecture. We use $5$ graph convolutional blocks, with $512$ output channels. This model is trained with an Adam optimizer, using a learning rate of $5e-2$.

We train this network using input the sequences formed via linear interpolation of the keypose values found within $125$ frames. These sequences are compared to their corresponding ground truth sequences. We train the network using three loss functions.
\vspace{-0.5em}\begin{itemize}
	\itemsep-0.2em
	\item $E_\text{pose}$: The MSE loss between the poses of the predicted sequence and the ground truth sequence.
	\item $E_\text{vel}$: The MSE loss between the velocities of the predicted sequence and the ground truth sequence.
	\item $E_\text{bone}$: The MSE loss between the bone lengths of the predicted sequence and the ground truth sequence.
\end{itemize}
The losses are combined as,
\begin{equation}
E = w_\text{pose} E_\text{pose} +w_\text{vel} E_\text{vel}+ w_\text{bone} E_\text{bone}\;,
\end{equation}
where $w_\text{pose}$, $w_\text{vel}$, and $w_\text{bone}$ weigh the different loss terms. We set these as $0.1, 100, 1e-6$ respectively, as these values give us the lowest validation $E_\text{pose}$ loss. We primarily consider this loss for validation as the other terms are more used as regularizers.

\subsection{Training-Time and Inference-Time Analysis} 
Keypose extraction using an error threshold of $500$ and k-means clustering using $1000$ clusters takes $0.2$ hours.

We compare the training time of different methods in Table~\ref{tab:training_time}. We trained the methods of TIM-GCN~\cite{Lebailly20} and HisRep~\cite{Mao20} for $50$ epochs, Mix\&Match~\cite{Aliakbarian20} f\label{key}or $100$k iterations, and DLow~\cite{Yuan20} for $500$ epochs each for both training steps, as recommended in their papers. We observe that our method is the fastest, with TIM-GCN also being relatively quick to train compared to the other methods. Note that our training time includes the keypose extraction and clustering, when this data is precomputed training takes $2.0$ hours. We have not included the training time for our refinement network, which takes an additional $0.6$ hours to pretrain, independently from the keypose prediction network.

Inference times for a single sequence is reported in Table~\ref{tab:inference_time}. We also include the keypose extraction time for a single sequence in our inference time. In this case, TIM-GCN gives the fastest results with $0.03$ seconds. Our method's inference time is $0.38$ seconds, which is sufficient for long-term predictions of $5$ seconds.

\begin{table}[!h]
\centering{
\begin{tabular}{|l|c|}
\hline                        &training-time (hours)  \\ \hline
TIM-GCN                 & 2.9    \\ \hline
HisRep10                    & 11.2   \\ \hline
HisRep125                  &  23.5 \\ \hline
Mix\&Match          & 14.8   \\ \hline
DLow          & 30.9   \\ \hline
Ours                        & \textbf{2.2}  \\ \hline
\end{tabular}}

\caption{{\bf Training-time of different methods}. Our method is the fastest to train among state of the art motion prediction methods.}
\label{tab:training_time}
\end{table}

\begin{table}[!h]
	\centering{
		\begin{tabular}{|l|c|}
			\hline                        &inference-time (seconds)  \\ \hline
			TIM-GCN                 & \textbf{0.03}    \\ \hline
			HisRep10                    & 0.22   \\ \hline
			HisRep125                  &  0.22 \\ \hline
			Mix\&Match          & 0.64   \\ \hline
			DLow          & 0.16   \\ \hline
			Ours                        & 0.38 \\ \hline
	\end{tabular}}
	
	\caption{{\bf Inference-time of different methods}. TIM-GCN is the fastest method for inference of a single sequence.}
	\label{tab:inference_time}
\end{table}

\subsection{Action Classifier Model}

One of the main metrics presented in our work is the motion-only action classifier (MOAC), partly based on HCN~\cite{Li19m}. We discuss this metric, as well as a Full Action Classifier (FAC) in this section.

\begin{figure}[!h]
	\centering
	\includegraphics[width=0.9\linewidth]{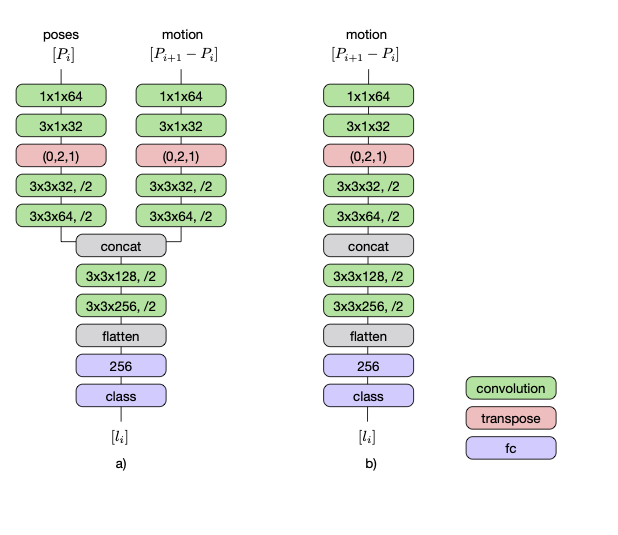}
	\caption{{\bf Action classifier architectures} based on~\cite{Li19m}. a) Full action classifier (FAC), which takes as input both the poses and the motion (difference between consecutive poses). b) Motion-only action classifier (MOAC), which takes only motion as input. The green blocks represent the convolution layers, where the three dimensions inside the block denote the kernel height, kernel width and the number of output channels respectively. A trailing ``/2" represents an appended MaxPooling layer with stride 2 after the convolution. The pink blocks represent transpose layers which transpose the dimensions of their input according to the order written on the block. The purple blocks represent the fully connected layers, with the number of output features of the layers.}
	\label{fig:ar_arch}
\end{figure}

\paragraph{Full Action Classifier (FAC) Model} In order to run a complete analysis, we report the results on a full action classifier (FAC). The full action classifier is based fully on HCN~\cite{Li19m}. As opposed to the only-motion action classifier (OMAC), it also processed the poses in a separate stream. The two architectures are depicted in Figure~\ref{fig:ar_arch}.

\paragraph{Results on FAC Metric}

We report the results of FAC on Human3.6M in Table~\ref{tab:fac}. We note the high performance of the mean-pose predictor on this metric. This predictor produces a sequence consisting of a single pose, which is set to the mean pose of the action category. The mean poses of the action categories were found using the training set. Despite having no motion whatsoever, this predictor is able to outperform everyone, except for the oracle, which evaluates the ground truth future motion. We also point out that the static predictor, which predicts a sequence consisting only of the last-seen pose, also does relatively well on this metric. This predictor also produces no motion whatsoever.

While HisRep125 performs best out of the prediction methods, it had a much lower MOAC performance compared to our method. This indicates that the motions we predict are more realistic than the ones HisRep125 produces but that the FAC classifier can use a few well-predicted static poses to guess the motion nevertheless. 

We also report the results of FAC on CMU Mocap Dataset in Table~\ref{tab:fac_cmu}. The trend is similar to the results on Human3.6M, with the extremely high performance of the mean-pose predictor. This further leads us to conclude that this metric is able to distinguish sequences which have no motion at all, therefore not being a reliable metric for our purposes.

\begin{table}[htbp]
\centering{
\begin{tabular}{|l|c|c|c|c|}
\hline                        &top-1 & top-2 & top-3 & top-5 \\ \hline
Oracle							 &  59   &  78    & 86  &94  \\ \Xhline{3\arrayrulewidth}
Mean-Pose							 &  \textbf{53} &  \textbf{67}    & \textbf{73}  & \textbf{80} \\  \hline
Static							 &  28   &  40    & 51  & 64 \\ \hline
TIM-GCN                 &  39 &  52  & 63  &76  \\ \hline
HisRep10                    & 36   & 53  & 63  & 76   \\ \hline
HisRep125                  &  42   & 56 & 67 & \textbf{80} \\ \hline
Mix\&Match                 &  29 & 45     & 58   & 71  \\ \hline
DLow                 &  23 & 46     & 54   & 65  \\ \hline
Ours                       & 38    & 50    & 60    & 72\\ \hline
\end{tabular}}

\caption{{\bf Results of the full action classifier (FAC) on the Human3.6M dataset}. We compare the classification accuracies of the motion predicted via our work compared to SOTA methods. The oracle evaluates ground truth future motions. Static evaluates a sequence consisting of only the last seen pose, i.e. not predicting any motion at all. Its relatively high performance despite being severely handicapped indicates that this is not a very reliable metric. Mean-pose evaluates a sequence consisting of a static pose set to the mean pose of the action category and is able to achieve results higher than everyone, emphasizing how little importance motion carries for this metric.}
\label{tab:fac}
\end{table}

\begin{table}[htbp]
	\centering{
		\begin{tabular}{|l|c|c|c|c|}
			\hline                        &top-1 & top-2 & top-3 & top-5 \\ \hline
			Oracle							 &  76   &  94    & 100  &100  \\ \Xhline{3\arrayrulewidth}
			Mean-Pose							 &  \textbf{100}   &  \textbf{100}    & \textbf{100}  & \textbf{100} \\  \hline
   			Static							 &  72   &  80    & 95  & \textbf{100}  \\ \hline
			TIM-GCN                 &  73 &  85  & 95  &\textbf{100}   \\ \hline
			HisRep10                    & 71   & 90  &  98  &\textbf{100}  \\ \hline
			HisRep125                  &  74   & 85  & 92     & \textbf{100} \\ \hline
			Mix\&Match                 &  44 & 51     & 66   & 94  \\ \hline
			Dlow             			  &  57 & 62     & 64   & 92  \\ \hline
			Ours                        & 73    & 79    & 88    & 96\\ \hline

	\end{tabular}}
	
	\caption{{\bf Results of the full action classifier (FAC) on the CMU-Mocap Dataset}. The trend is similar to the results on Human3.6M. In this case, it is even more striking that the Mean-Pose predictor is able to get 100\% accuracy on even top-1, with using just a single static pose. Static, which evaluates a sequence consisting of only the last seen pose also has a high performance rivalling the results of the learning-based prediction methods.}
	\label{tab:fac_cmu}
\end{table}

\paragraph{Training Details} The action recognition models are trained to classify sequences of $125$ frames ($5$ seconds). Both models are trained with Adam optimizer, using a learning rate of $1e-5$, and weight decay of $1e-4$. 
In order to make the model more robust to overfitting, we have added Gaussian noise during training to the input data. With independent probabilities of $0.5$, a noise of $20$ mm standard deviation is added to the motion and noise of $30$ mm standard deviation is added to the poses. We have found that this procedure improves the validation accuracies.
The trained models and the training code will also be released by us upon acceptance.

We note that we have tried to base our action recognition model on more recent work, such as SGN~\cite{Zhang20e}. This method shows accuracies above 90\% on NTU-D dataset, which has shorter and very distinct actions, and not on $5$ second H36M sequences. Nevertheless, we have trained SGN on our sequences as a motion-only classifier and the accuracies it achieves on ground truth sequences are: $50, 68, 76, 85$ corresponding to top-1, top-2, top-3, and top-5 accuracies respectively. This is slightly worse than the classifier based on HCN, therefore we have chosen to continue using an HCN based model.

\subsection{Cluster center visualization} 
\begin{figure*}[!h]
	\centering
	\includegraphics[width=0.9\linewidth]{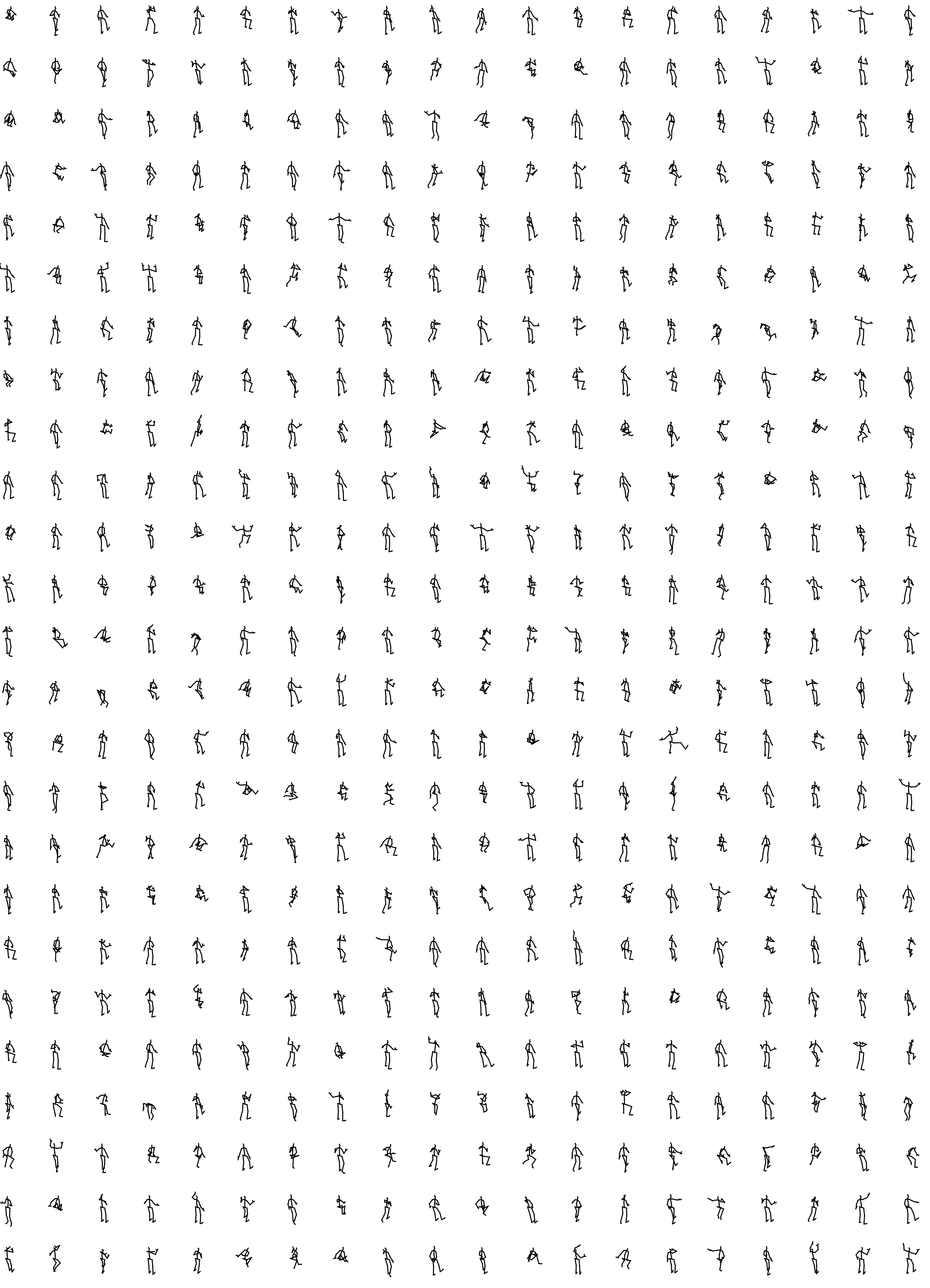}
		\caption{{\bf Visualization of keypose cluster centers}. The sampled $500$ keypose cluster centers here show that the cluster centers are quite varied and are able to represent the keyposes throughout the different categories of motions.}
	\label{fig:cc}
\end{figure*}

We show a sample of $500$ keypose cluster centers in Figure~\ref{fig:cc}. It is necessary for them to be varied in order to be able to express the wide range of poses seen across different action categories. We find that the keypose cluster centers include poses from many categories, such as sitting, crouching, squatting, standing, walking, and making different arm gestures.

\subsection{Additional Qualitative Results}
\begin{figure*}[h!]
	\centering
	\includegraphics[width=1.0\linewidth]{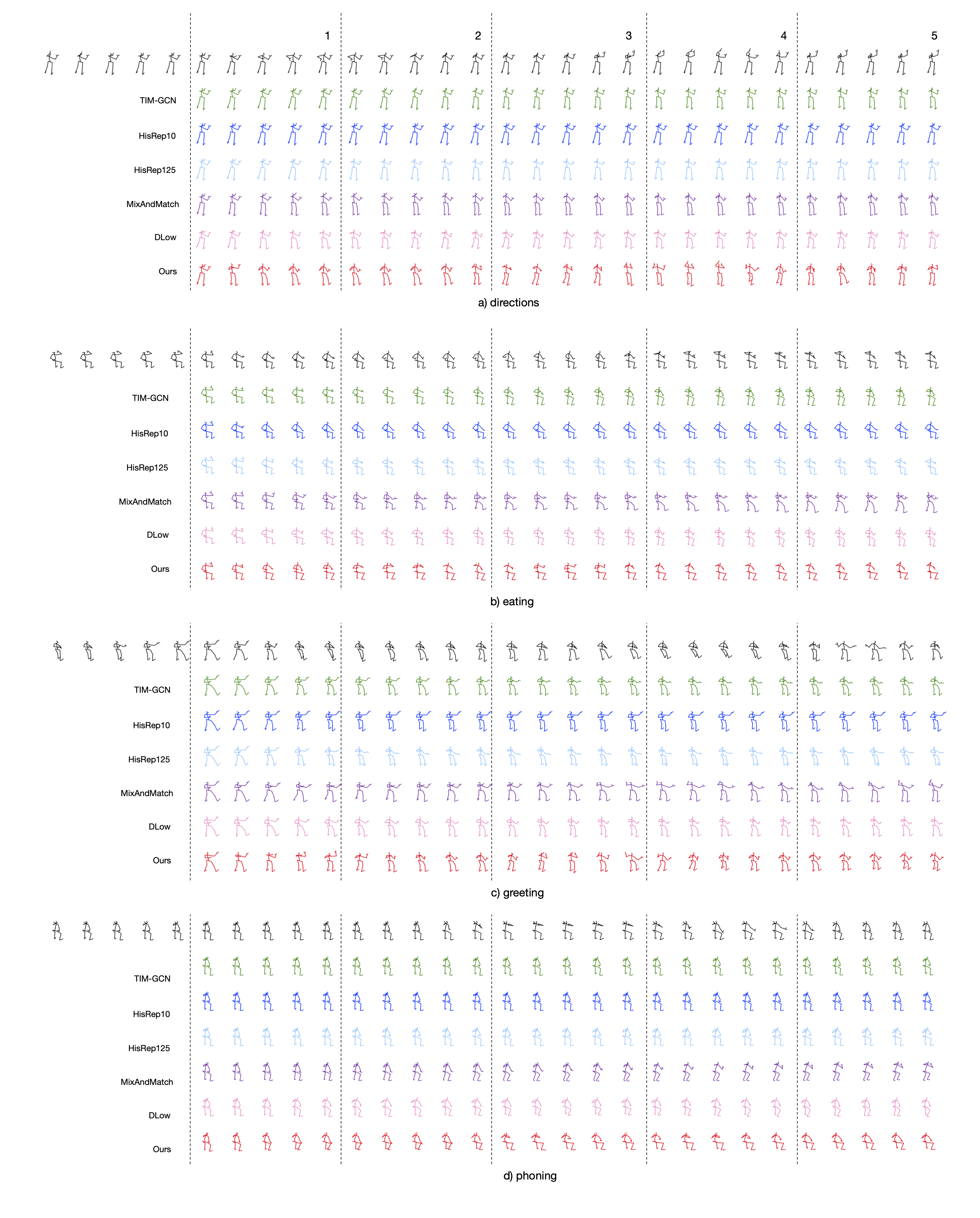}
		\caption{{\bf Qualitative results on Human3.6M} of actions ``directions",``eating", ``greeting", ``phoning".}
	\label{fig:qual1}
\end{figure*}

\begin{figure*}[h!]
	\centering
	\includegraphics[width=1.0\linewidth]{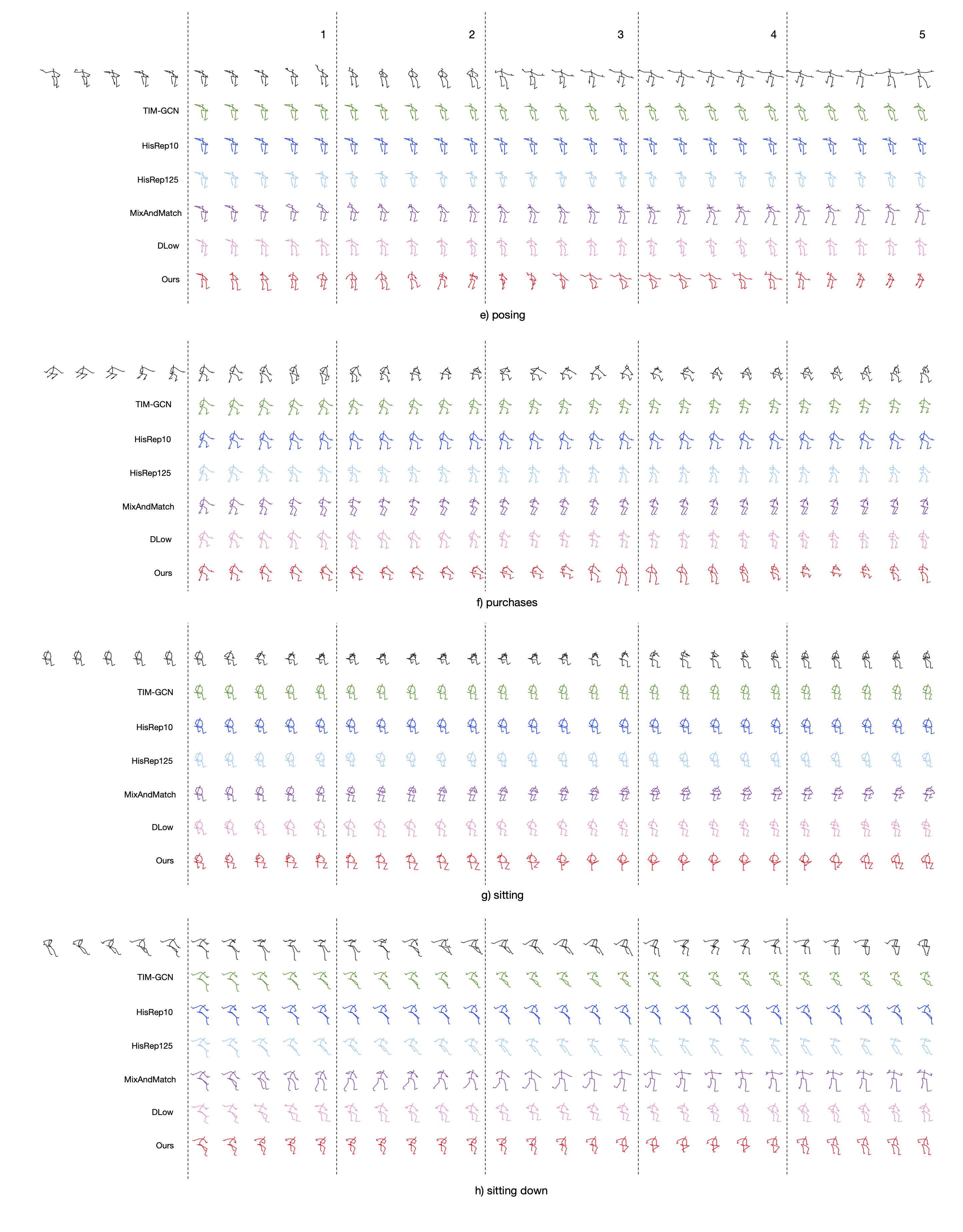}
		\caption{{\bf Qualitative results on Human3.6M} of actions ``posing", ``purchases", ``sitting", ``sitting down".}
	\label{fig:qual2}
\end{figure*}

\begin{figure*}[h]
	\centering
	\includegraphics[width=1.0\linewidth]{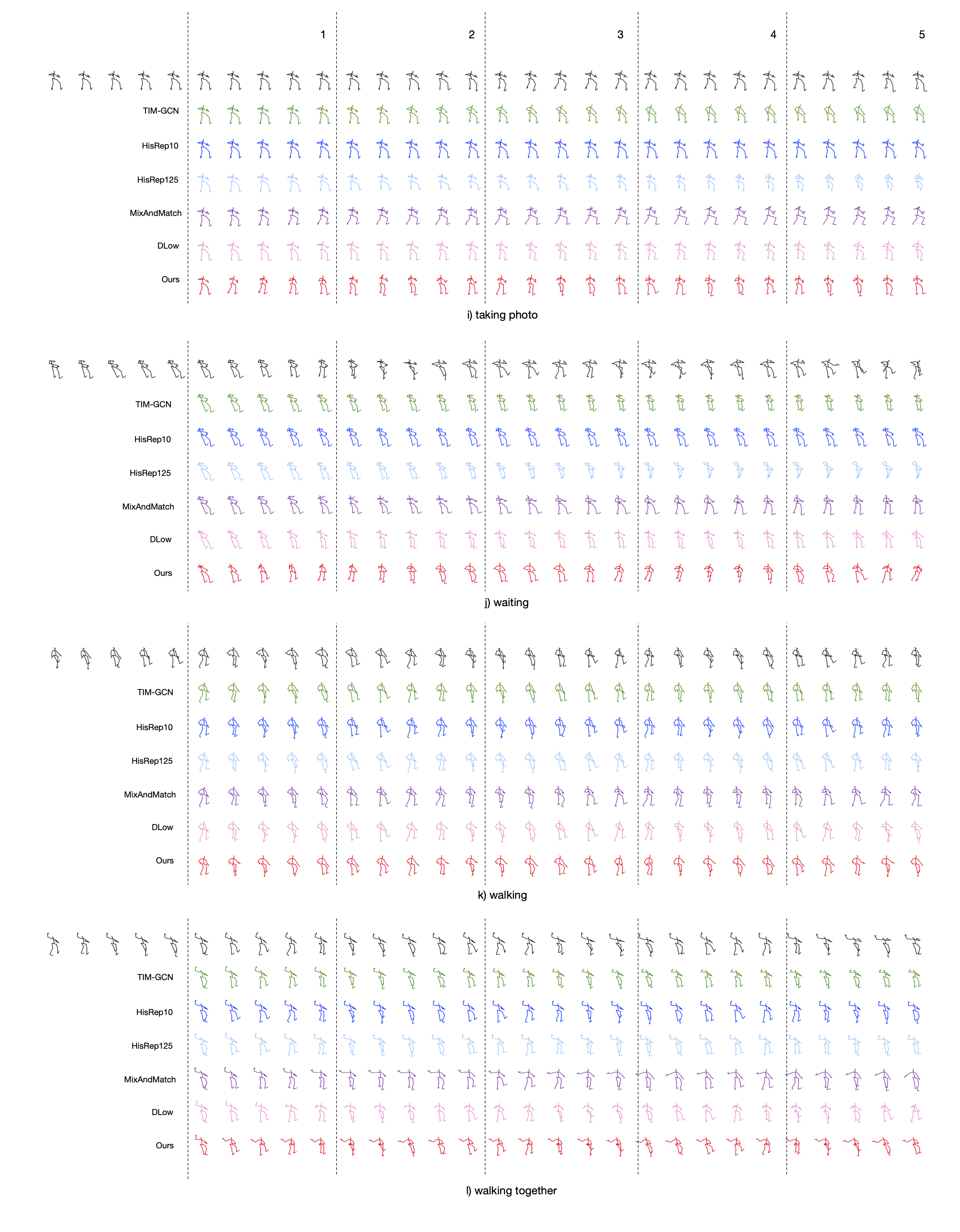}
		\caption{{\bf Qualitative results on Human3.6M} of actions ``taking photo", ``waiting", ``walking", ``walking together".}	\label{fig:qual3}
\end{figure*}

Additional qualitative results on the Human3.6M dataset can be seen in Figures~\ref{fig:qual1},\ref{fig:qual2},\ref{fig:qual3}. The top row in black depicts the ground truth poses and the first five poses represent the last seen $1$ second of the conditioning ground truth. We display the action categories which we were unable to show in our main paper due to lack of space. We again draw attention to the wide gestures our model is able to generate and the overall dynamism of the predicted motions as compared to the SOTA methods.

\begin{figure*}[h!]
	\centering
	\includegraphics[width=1.0\linewidth]{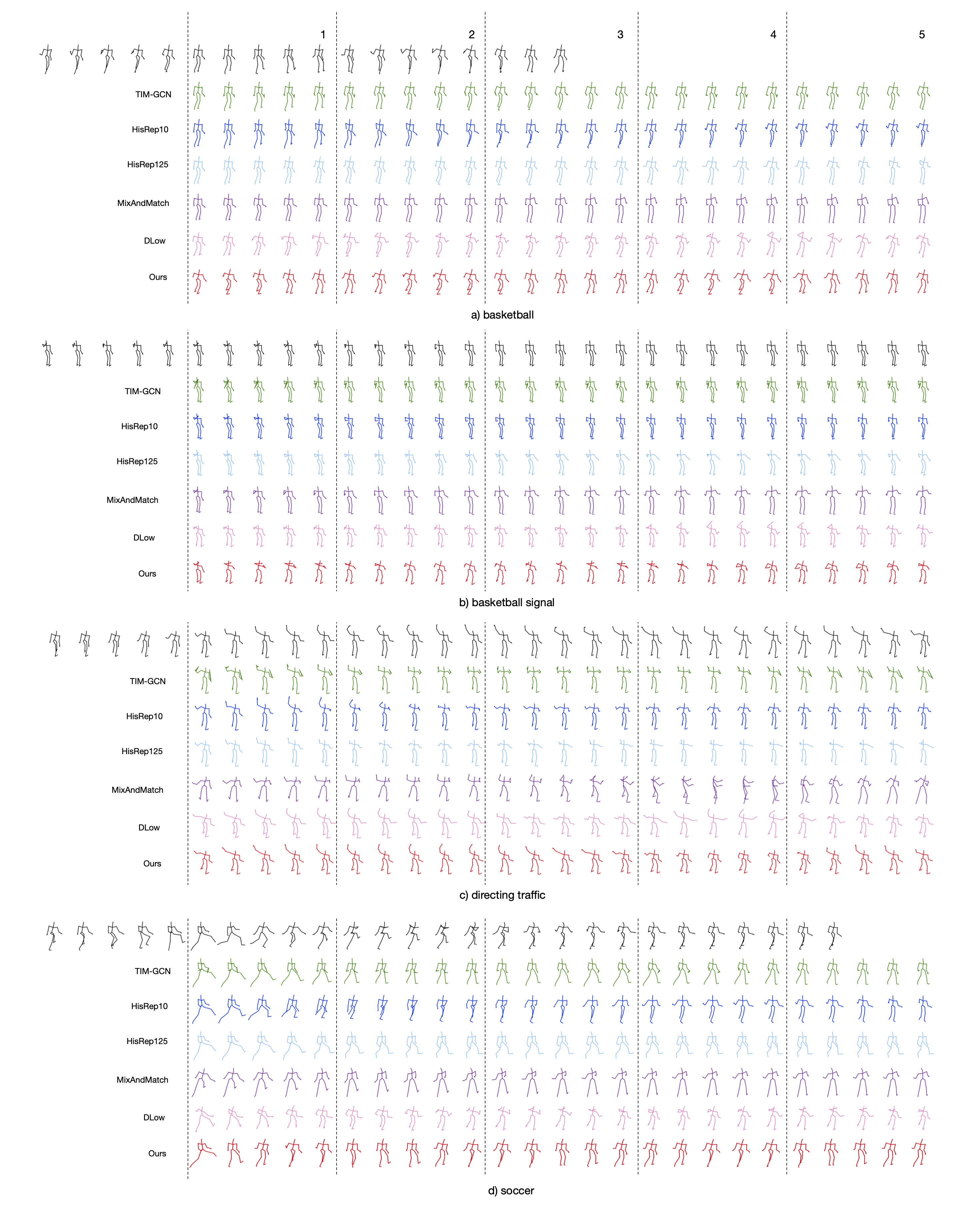}
		\caption{{\bf Qualitative results on CMU Mocap} of actions ``basketball",``basketball signal", ``directing traffic", ``soccer".}

	\label{fig:qual1_cmu}
\end{figure*}

\begin{figure*}[h!]
	\centering
	\includegraphics[width=1.0\linewidth]{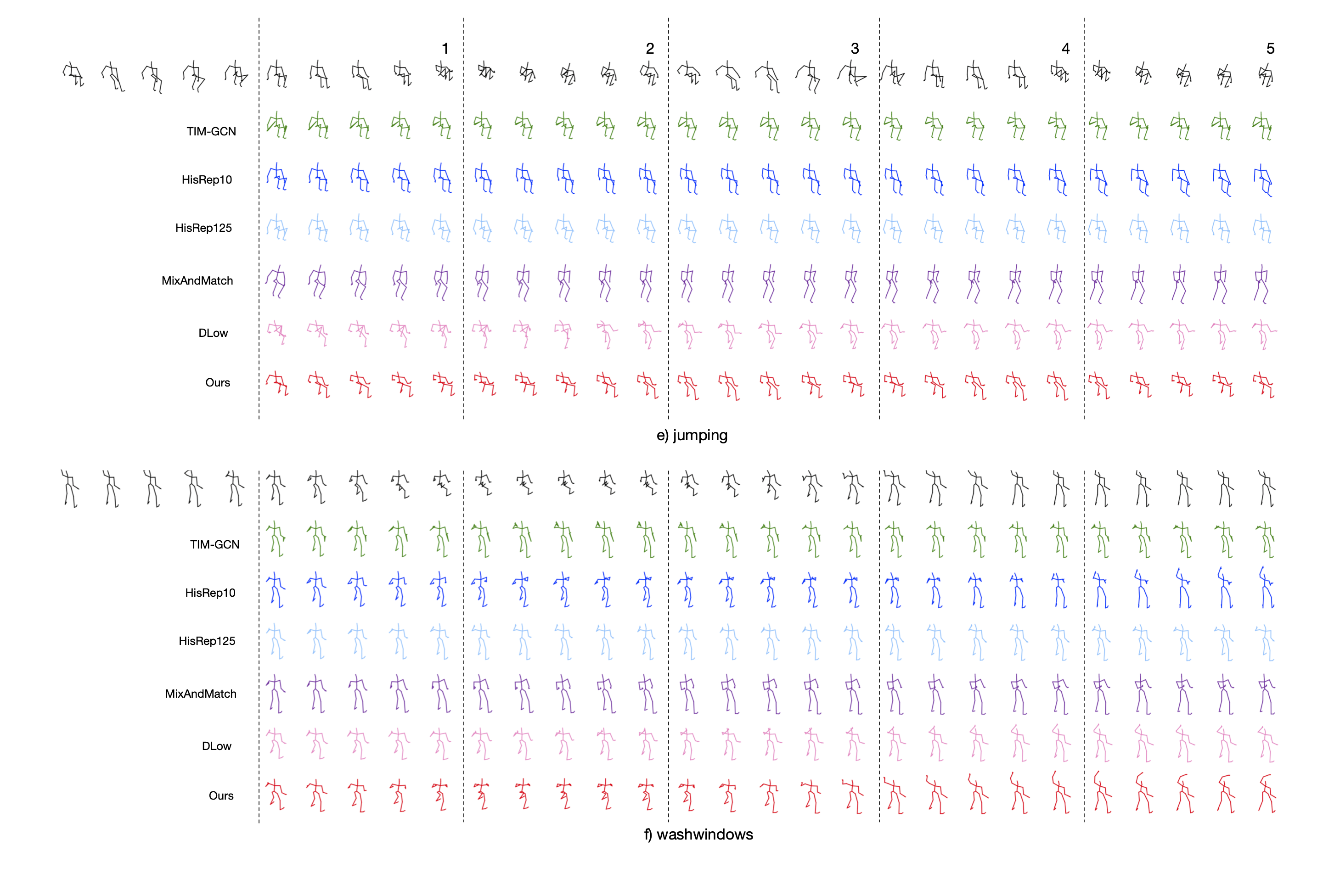}
		\caption{{\bf Qualitative results on CMU Mocap} of actions ``jumping",``washwindows".}

	\label{fig:qual2_cmu}
\end{figure*}

Qualitative results for the CMU-Mocap dataset can be seen in Figures~\ref{fig:qual1_cmu},\ref{fig:qual2_cmu}. We see that we achieve more dynamic and realistic poses on this dataset as well.

\end{document}